\documentclass[runningheads]{llncs}
\usepackage[T1]{fontenc}
\usepackage{graphicx}
\usepackage{amsmath}
\usepackage{booktabs}
\usepackage{tabularx}
\usepackage[misc,geometry]{ifsym}
\usepackage{cite}
\usepackage{hyperref}
\makeatletter
\newcommand{\printfnsymbol}[1]{%
  \textsuperscript{\@fnsymbol{#1}}%
}
\makeatother
\begin{document}
\title{Fusionista2.0: Efficiency Retrieval System for Large-Scale Datasets}
\author{Huy M. Le\inst{1, 2,3,5} \and
Dat Tien Nguyen\inst{1, 2} \and
Phuc Binh Nguyen\inst{3,5}  \and Gia Bao Le Tran\inst{3,5} \and Phu Truong Thien\inst{3,5} \and
Cuong Dinh\inst{3,5} \and Minh Nguyen\inst{3,5} \and
Nga Nguyen\inst{3,5}
\and Thuy T. N. Nguyen\inst{3,5} \and
Tan Nhat Nguyen\inst{3,5} \and
Binh T. Nguyen\inst{2,4,5} }
% \orcidID{0009-0002-4767-5382}
% \orcidID{0009-0004-4606-3610}
% \orcidID{0009-0006-8370-694X}
% \orcidID{0009-0005-2837-3145}
% \orcidID{0009-0000-4273-1651}
%
\authorrunning{Huy et al.}
\institute{Mohamed Bin Zayed University of Artificial Intelligence (MBZUAI), UAE \and AISIA Research Lab, Ho Chi Minh City \and University of Information Technology, Ho Chi Minh City, Vietnam \and Ho Chi Minh University of Science, Vietnam \and Vietnam National University, Ho chi Minh City, Vietnam   \\
\email{HuyM.Le@mbzuai.ac.ae}}
\maketitle              % typeset the header of the contribution
\begin{abstract}
The Video Browser Showdown (VBS) challenges systems to deliver accurate results under strict time constraints. To meet this demand, we present \texttt{Fusionista2.0}, a streamlined video retrieval system optimized for speed and usability. All core modules were re-engineered for efficiency: preprocessing now relies on ffmpeg for fast keyframe extraction, Optical Character Recognition (OCR) is powered by Vintern-1B-v3.5 for robust multilingual text recognition, and Automatic Speech Recognition (ASR) employs faster-whisper for real-time transcription. For question answering, lightweight vision–language models provide quick responses without the heavy cost of large models. Beyond these technical upgrades, \texttt{Fusionista2.0} introduces a redesigned UI/UX with improved responsiveness, accessibility, and workflow efficiency, enabling even non-expert users to retrieve relevant content rapidly. Evaluations demonstrate that retrieval time was reduced by up to 75\% while accuracy and user satisfaction both increased, confirming \texttt{Fusionista2.0} as a competitive and user-friendly system for large-scale video search.

\keywords{Multimedia Retrieval  \and Efficient System \and Multimodal.}
\end{abstract}
\section{Introduction}
As noted in prior works~\cite{Manning2008IR}, effective information retrieval not only improves user experience but also supports decision-making processes across a wide range of domains. Consequently, research efforts must focus on developing retrieval methodologies with superior efficacy or refining and optimizing existing approaches to enhance their scalability within this domain and foster seamless interaction between the retrieval system and the user. Addressing these points, Video Browser Showdown~\cite{schoeffmann2012video}, is an established, biennial competition that serves as a robust benchmark for assessing state-of-the-art interactive video retrieval systems. The VBS operates within realistic constraints, encompassing a spectrum of complex ad-hoc search issues, specifically: (\textit{1}) Known-Item Search (KIS), (\textit{2}) Ad-hoc Video Search (AVS), and (\textit{3}) Visual Question Answering (VQA). 

For VBS 2026, the primary dataset is the V3C~\cite{rossetto2019v3c}, comprising over 28,000 videos with thousands of hours of content. Additional domain-specific datasets, such as Marine Video~\cite{marinevideo} and LapGynLHE~\cite{lapgynlhe}, are also included. These heterogeneous datasets pose unique challenges and provide a comprehensive benchmark for testing the robustness and adaptability of retrieval systems.

In this paper, we present \texttt{Fusionista2.0}, an enhanced information retrieval system that extends \texttt{Fusionista}~\cite{fusionista} and \texttt{Fustar}~\cite{fustar} with multiple modalities, including text-to-image, image-to-image, object-level, and OCR/ASR-based retrieval. The main improvements are: 
\begin{itemize}
    \item[(\textit{a})] A redesigned UI/UX that enhances accessibility for non-expert users and reduces the learning curve.
    \item[(\textit{b})] Optimized data processing and search mechanisms for faster retrieval.
    \item[(\textit{c})]A powerful re-ranking module to refine results and maximize accuracy.
\end{itemize}
Statistics evaluations with new users show that \texttt{Fusionista2.0} have significantly improve in both efficiency and effectiveness, advancing interactive video retrieval while lowering the entry barrier for broader real-world use.

\section{Key functions upgrade}
\subsection{Overview}
\vspace{-1em}
\begin{figure}
    \centering
\includegraphics[width=\linewidth]{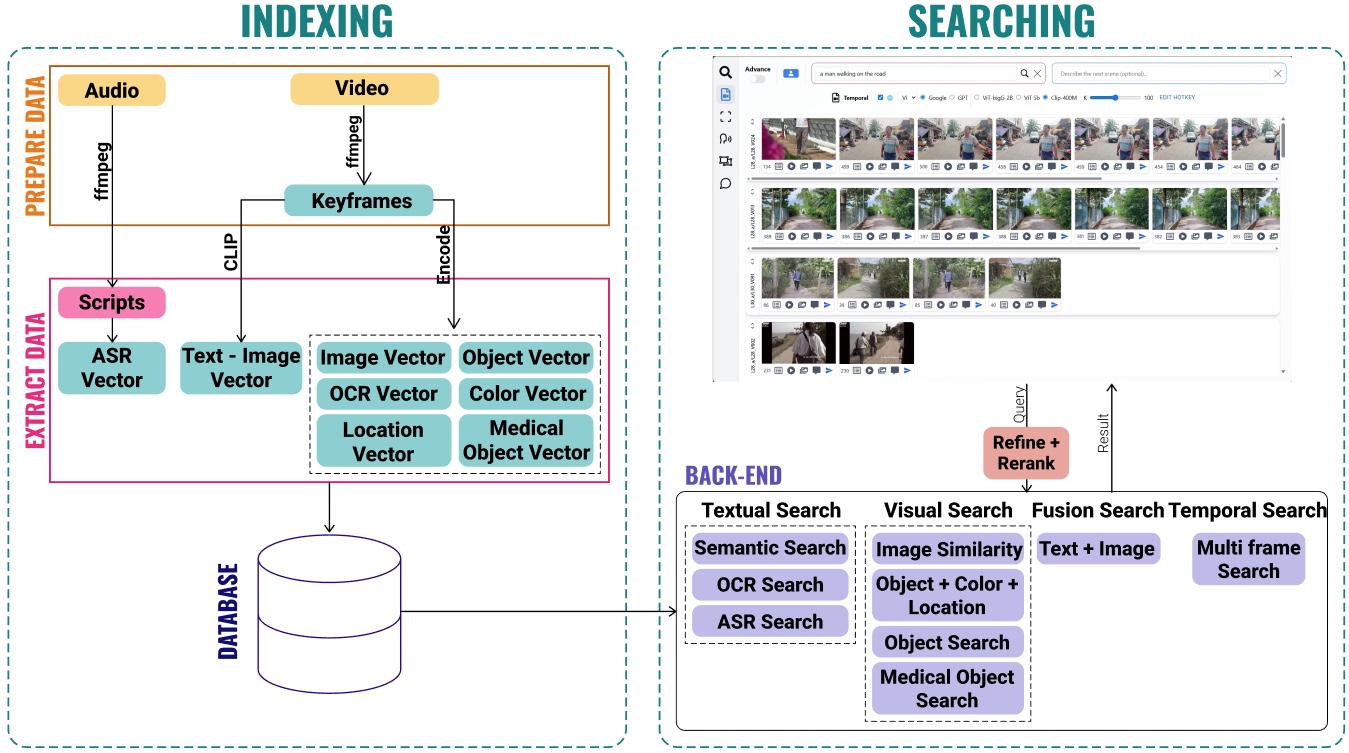}
    \caption{Overview of out system.}
    \label{fig:overview}
    \vspace{-2em}
\end{figure}
Our system, \texttt{Fusionista2.0}, builds upon the design principles and techniques of its predecessors, \texttt{Fusionista}~\cite{fusionista} and \texttt{Fustar}~\cite{fustar}. It provides a comprehensive set of retrieval functionalities tailored for the diverse query types in VBS2025. While these functionalities proved effective in VBS2025, scaling the system to multi-terabyte datasets revealed several challenges such as inference time, UX flow for thousand images. In \texttt{Fusionista2.0}, we introduce six key improvements (Figure~\ref{fig:overview}), including:
\begin{enumerate}
    \item Optimized Data Preparation: The new all-in-one preprocessing pipeline is highly optimized for video decoding, frame extraction, and keyframe selection. This ensures faster, resource-efficient processing of large-scale datasets. 
    \item Enhanced Textual Search: Multiple embedding models are combined into an ensemble to improve semantic coverage and accuracy.  
    \item Efficient OCR/ASR: Large OCR and ASR models are replaced by lightweight alternatives with comparable accuracy, ensuring scalability.  
    \item Modern VLLM-based Question Answering: The QA module is upgraded with recent Vision–Language Large Models (VLLMs), optimized for both reasoning and inference speed, delivering faster and accurate answers.  
    \item Reranking with Interactive Confirmation: A novel reranking mechanism leverages clarification questions to refine the search results according to user intent, reducing the risk of missing subtle details.   
    \item UI/UX Enhanced: As user interaction is the most critical aspect in interactive retrieval, we redesigned the interface to minimize redundant browsing of repeated video content, aligning the layout with natural user viewing flow.  
\end{enumerate}  
Through these improvements, \texttt{Fusionista2.0} achieves a balanced trade-off between retrieval effectiveness, efficiency, and user interactivity, addressing the challenges of large-scale multimedia search in VBS2026.  
\vspace{-0.5em}
\subsection{Data Preparation and Extraction}
In our previous system, the preprocessing pipeline was designed to maximize the accuracy of keyframe selection by leveraging multiple state-of-the-art models. Our previous system employed a preprocessing pipeline that combined CLIP-B/32 embeddings~\cite{radford2021}, TransNetV2~\cite{souvcek2020} scene detection, and clustering to select representative keyframes~\cite{tan2021}, which were then used for multi-function search.

Although this method achieved high precision in selecting representative keyframes, it was computationally intensive, requiring significant GPU and memory resources, and thus proved inefficient for large-scale datasets like V3C.

To overcome scalability issues, the multi-model pipeline is replaced with an all-in-one \textit{ffmpeg}-based keyframe extraction workflow. The system deterministically analyzes each video stream and isolates all intra-coded frames that serve as structural keyframes, exporting them as sequential images along with a precise timestamp–frame mapping for reproducibility. This streamlined approach greatly reduces computational and memory demands, enabling efficient large-scale preprocessing while maintaining sufficient quality for frame-extraction tasks, thus ensuring feasibility under the competition's constraints.
\vspace{-0.5em}
\subsection{Textual Search}
In our previous system, text-to-image retrieval relied on a single CLIP-based model, which often required a trade-off between efficiency and accuracy. In \texttt{Fusionista2.0}, we enhance this module by employing two state-of-the-art CLIP variants~\cite{clipbench}: \textit{CLIP-Sig400M} and \textit{CLIP-ViT-5B}. Both models are widely recognized for their balanced performance in terms of inference speed and retrieval accuracy, making them well-suited for large-scale interactive search scenarios.  

\begin{equation}
s(q,v) = \alpha \cdot s_{\text{Sig400M}}(q,v) + (1 - \alpha) \cdot s_{\text{ViT-5B}}(q,v)
\end{equation}
The value of $\alpha$ is fixed equal to 0.7, which is empirically chosen based on extensive internal experimentation and a small-scale user study (Table~\ref{tab:choosealpha}) conducted with 50 participants interacting with our system to find best result for a text-image retrieval question. This adaptive weighting allows \texttt{Fusionista2.0} to capture semantic nuances more robustly than using either model in isolation.
\vspace{-1em}
\begin{table}[ht]
\centering
\caption{Number of people can retrieve the result at top-1 with different $\alpha$ thresholds and model backbones on the person-matching task.}
\label{tab:choosealpha}
\begin{tabularx}{\linewidth}{|l|*{9}{>{\centering\arraybackslash}X|}}
\hline
$\alpha$       & 0.1 & 0.2 & 0.3 & 0.4 & 0.5 & 0.6 & 0.7 & 0.8 & 0.9 \\ \hline
Number of People     & 34  & 35  & 31  & 40  & 39  & 38  & 43  & 29  & 38  \\ \hline
CLIP Sig400M   & \multicolumn{9}{c|}{38} \\ \hline
CLIP ViT-5B    & \multicolumn{9}{c|}{33} \\ \hline
\end{tabularx}
\end{table}
\vspace{-3em}
\subsection{OCR and ASR}
In \texttt{Fusionista2.0}, we replace the previous PaddleOCR pipeline with the \textit{Vintern-1B-v3.5} model ~\cite{vintern}. This model, fine-tuned from InternVL2.5-1B~\cite{intervl}, leverages low-resource languages datasets such as Viet-ShareGPT-4o-Text-VQA~\cite{vintern} and over one million OCR samples, achieving strong performance in many languages. Vintern-1B-v3.5 not only delivers high accuracy in scene text detection and recognition but also benefits from enhanced reasoning and general knowledge, enabling it to infer characters that are blurred or occluded.

 While Whisper~\cite{whisper} provides excellent accuracy, its large model size makes it impractical for large-scale processing. Moreover, the majority of audio tracks in the VBS datasets contain ambient sounds rather than speech, reducing the necessity for high-capacity ASR models. For these reasons, we replace Whisper with \textit{faster\_whisper}~\cite{faster-whisper}, a lightweight yet efficient variant optimized for speed. This choice significantly accelerates the transcription pipeline by four times~\cite{faster-whisper}, while still maintaining sufficient accuracy to extract meaningful speech content. 
 \vspace{-2em}
\subsection{Question Answering}
\vspace{-1.5em}
\begin{table}[ht]
\scriptsize
\centering
\caption{Summary performance per model.}
\label{tab:qamodel}
\begin{tabular}{lcccc}
\toprule
Model & Count Acc. & Img. Ans. Acc. & Video Anss Acc. & Avg. Time (s) \\
\midrule
InternVL-1B-Seq~\cite{intervl} & 0.72 & 0.68 & 0.67 & 2.52 \\
InternVL-1B-ffn6~\cite{intervl} & 0.79  & 0.74 & 0.71 & 2.48 \\
InternVL-1B-ffn6-Seq~\cite{intervl} & \textbf{0.84} & \textbf{0.79} & \textbf{0.75} & 4.80 \\
LLaVA-0.5B-ffn6~\cite{llava} & 0.78  & 0.63 & 0.73 & 10.50 \\
SmolVLM-0.5B-ffn6~\cite{smovlm} & 0.83 & 0.52 & 0.65 & \textbf{1.07} \\
\bottomrule
\vspace{-2em}
\end{tabular}
\end{table}

While large vision-language models (VLMs, $\geq$7B) excel in video question answering, they are too slow and often less accurate than humans, making them impractical for time-critical retrieval tasks like VBS. Our approach prioritizes speed and reliability by targeting simple but frequent queries—such as object counting, attribute recognition, and text extraction—while offering partial support for human for complex reasoning tasks. This strategy ensures fast responsiveness while keeping human expertise in the loop where models still fall short.

To systematically choose the right lightweight models, we constructed a benchmark dataset of 200 video–question pairs from previous queries of VBS, categorized into three main groups: (1) Counting (\textit{How many completed shoes in the image?}), (2) Image information extraction (\textit{What is on the street?}), and (3) Video information extraction (\textit{What color is the phone the woman is using?}). The dataset requires models to provide an image/video. We then tested several compact VLMs ($\leq$1B parameters) on this dataset, with the aim of identifying the best compromise between accuracy and inference speed. The detailed results are summarized in Table~\ref{tab:qamodel}.  

Among all candidates, the \textit{InternVL-1B-ffn6-Seq}~\cite{intervl} model demonstrated the most consistent performance across the three categories. It achieved an average inference time of less than 5 seconds per query over the entire dataset, while maintaining high accuracy in counting and frame-level information extraction tasks. This balance of speed and reliability makes it the most suitable choice for our system, ensuring accurate answers within the strict time limits of VBS.  
\vspace{-1em}
\subsection{Reranking}
\vspace{-2em}
% \textbf{TO DO: cite the work and add image}

\begin{figure}
    \centering
    \includegraphics[width=\linewidth]{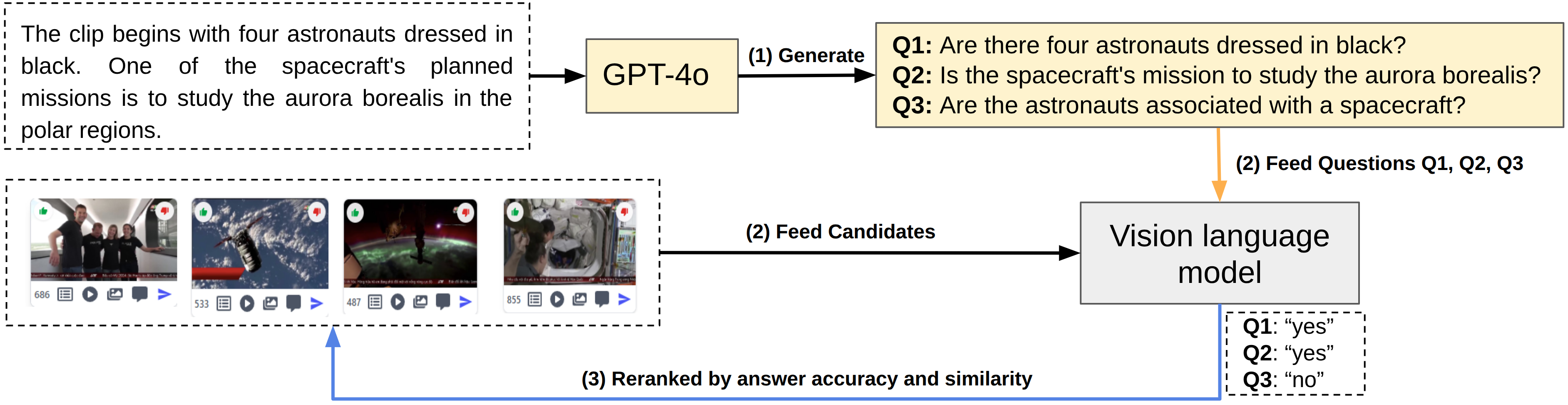}
    \caption{Illustration of reranking process. Images from prior search is reranked based on number of yes answers from the Vision language model}
    \label{fig:rerank}
    \vspace{-1em}
\end{figure}

Our reranking approach is inspired by the methodology outlined in \cite{aaaivqa4cir, summary}, which utilizes a post-processing strategy that can be effortlessly incorporated into existing image retrieval systems. However, we extend this concept by designing a fully integrated AI-based VQA pipeline. In this pipeline, we employ GPT-4o \cite{openai2024gpt4ocard} to formulate three yes-no questions based on the initial user query. These questions are crafted to explore the details of objects mentioned in the query, such as "\textit{Is there a dog in the scene?}" or "\textit{Is the dog colored yellow?}" Subsequently, both the questions and the associated images are evaluated by a vision language model to generate accurate responses. To enhance flexibility, we test multiple vision language models, including VideoLLaMA \cite{DBLP:conf/emnlp/ZhangLB23_video_llama} and BLIP-2 \cite{10.5555/3618408.3619222}, leveraging the VLLMs\cite{kwon2023efficient_vllm_lib} library to enable API-driven interactions.
\vspace{-0.5em}

\section{UI/UX Upgrade}
\vspace{-2em}
\begin{figure}
    \centering
    \includegraphics[width=\linewidth]{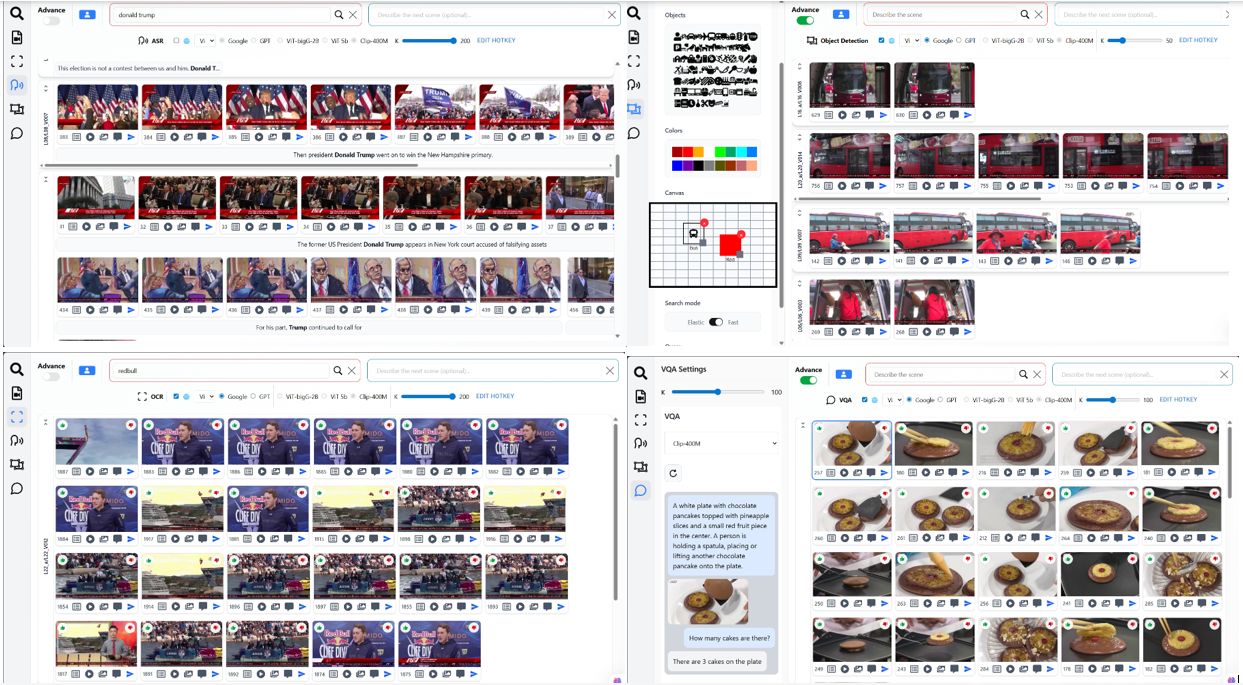}
    \caption{Our system's UI/UX for main functions. On the top-left, top-right, bottom-left, bottom-right is ASR search, object search, OCR search, QA respectively.}
    \label{fig:mainfunction}
\end{figure}
\vspace{-1em}
Our main key focus is UI/UX optimization, as efficient interaction directly impacts retrieval speed and usability. We redesigned the interface for responsiveness, accessibility (\textit{WCAG}~\cite{WCAG21} compliance with \textit{shadcn/ui})~\cite{shadcn}, better feedback via loading states and error handling. Migration from \textit{Create React App (CRA)}~\cite{cra} to \textit{Vite}~\cite{vite} enhance significantly system performance.

Workflows were streamlined with grouped search results, sidebar navigation with shortcuts, virtual scrolling, and multilingual query support. A conversational VQA interface and batch operations further boosted efficiency. These changes had a measurable impact on user interactions (showed in Figure~\ref{fig:mainfunction}. Overall, the optimizations greatly enhanced responsiveness, accuracy, and user satisfaction, ensuring \texttt{Fusionista2.0}’s competitiveness in time-critical video search.
\vspace{-2em} 
\section{Conclusion}
\vspace{-0.75em}
\texttt{Fusionista2.0} demonstrates that optimizing speed across all modules and redesigning the UI/UX can significantly improve both efficiency and usability in large-scale video retrieval. These advances make the system highly competitive for VBS2026 while lowering the barrier for real-world adoption.
\vspace{-0.5em}
\begin{credits}
\subsubsection{\ackname} This research was supported by The Mohamed bin Zayed University of Artificial Intelligence's Scientific Research Fund and The VNUHCM-University of Information Technology's Scientific Research Support Fund.
\end{credits}

\bibliography{bibliography}
\bibliographystyle{splncs04}
\end{document}